\documentclass{article}

\usepackage[utf8]{inputenc} 
\usepackage[T1]{fontenc}    
\usepackage{hyperref}       
\usepackage{url}            
\usepackage{booktabs}       
\usepackage{amsfonts}       
\usepackage{nicefrac}       
\usepackage{microtype}      
\usepackage{times}
\usepackage{graphicx} 
\usepackage{subfigure}

\usepackage{algorithm}
\usepackage{algorithmic}

\usepackage{geometry}
\usepackage{booktabs}

\usepackage{algorithm,algorithmic}

\usepackage{hyperref}

\usepackage{relsize}

\usepackage{bigints}

\usepackage{graphicx}

\usepackage{caption}

\usepackage{multirow}

\usepackage{float}

\usepackage{array}

\usepackage{setspace}
\newlength\iwidth
\newlength\iheight

\usepackage{paralist}

\usepackage{amsmath}
\usepackage{amssymb}
\usepackage{amsfonts}
\usepackage{amsthm}

\usepackage{tikz} \usetikzlibrary {plotmarks}
\usepackage{pgfplots}

\newlength{\myrowheight}
\newcommand {\R}   {{\rm I\!R}}

\newcommand {\bftheta} { { \boldsymbol \theta} }

\newcommand{\bfK}{{\bf K}}

\newcommand{\bfx}{{\bf  x}}
\newcommand{\bfy}{{\bf  y}}

\newcommand{\bfc}{{\bf c}}
\renewcommand{\bfK}{{\bf K}}

\newcommand{\bfP}{{\bf P}}

\newcommand{\bfR}{{\bf R}}

\newcommand{\bfL}{{\bf L}}

\newcommand{\bfI}{{\bf I}}
\newcommand{\bfW}{{\bf W}}

\newcommand{\bfs}{{\bf s}}

\newcommand{\bfb}{{\bf b}}

\newtheorem{example}{Example}

\usepackage{mathtools}
\usepackage{url}

\graphicspath{{./img/}}

\title{Learning Across Scales - Multiscale Methods for Convolution Neural Networks}

\author{ Eldad Haber\thanks{Department of Earth and Ocean Science, The University of British Columbia, Vancouver, BC,
Canada \texttt{eldadhaber@gmail.com}},\; Lars Ruthotto\thanks{Department of Mathematics and Computer Science, Emory University, Atlanta, GA, USA, \texttt{lruthotto@emory.edu}},\;  Elliot Holtham\thanks{Xtract Technologies, Vancouver, BC, Canada, \texttt{elliot@xtract.tech}},\;  Seong-Hwan Jun\thanks{Department of Statistics, The University of British Columbia, Vancouver, BC, Canada \texttt{seong.jun@stat.ubc.ca}}}

\begin{document}

\maketitle

\begin{abstract}
In this work we establish the relation between optimal control and training deep Convolution Neural Networks (CNNs). 
We show that the forward propagation in CNNs can be interpreted as a time-dependent nonlinear differential equation and learning as controlling the parameters of the differential equation such that the network approximates the data-label relation for given training data. Using this continuous interpretation we derive two new methods to scale CNNs with respect to two different dimensions. The first class of multiscale methods connects low-resolution and high-resolution data through prolongation and restriction of CNN parameters. We demonstrate that this enables classifying high-resolution images using CNNs trained with low-resolution images and vice versa and warm-starting the learning process.  The second class of multiscale methods connects shallow and deep networks and leads to new training strategies that gradually increase the depths of the CNN while re-using parameters for initializations. 
\end{abstract}

\section{Introduction}

In this work we consider the problem of designing and training Convolutional Neural Networks (CNNs). The topic has been a major field of research over the last years, after it has shown remarkable success, e.g., in classifying images of hand writing, natural images, videos; see, e.g., ~\cite{LeCunBengio1995,KrizhevskySutskeverHinton2012,LeCunKavukcuogluFarabet2010}
and references within. This success has generated thousands of research papers and a few celebrated software packages.

However, the success of CNNs is not fully understood and in fact, tuning network architecture and parameters is very hard in practice.
Typically many trial and error experiments are required to find a CNN that is effective for a specific class of data. 
In addition to the computational costs associated with those experiments, in many cases, small changes to the network can yield large changes in the learning performance.
To overcome the difficulty of learning, systematic approaches using Bayesian optimization have recently been proposed to infer the best architecture~\cite{GeneralFrameworkBayesOpt2016}.

Currently used training methods depend, e.g., on the architecture of the network as well as the resolution of the image data.
Changing any of those parameters in the training or prediction phase can severely affect the performance of the CNN.
For example, CNNs are typically trained using images of a fixed resolution, and classifying images of a different resolution requires interpolation.
Such a process can be computationally expensive, particularly if the data represents videos or high-resolution 3D images as is common in applications, e.g., in  medical imaging and geosciences~\cite{JiangTrundleRen2010,KarpathyCVPR14}).

In this paper we derive a framework that allows scaling CNNs across image resolution and depths and thus enables multiscale learning. 
As a backbone of our methods we present an interpretation of deep CNNs as an optimal control problem involving a nonlinear time dependent differential equations. This understanding leads to a very common structure that is used in fields such as path planning, data assimilation, and nonlinear Kalman filtering; see~\cite{PDEOptBook} and reference within.

We present new methods for scaling CNNs from low- to high-resolution image data and vice versa. 
We propose an algebraic multigrid approach to adapt the coefficients of the convolution kernel for different scales and demonstrate the importance of this step.
Our method allows multiscale learning using image pyramids, where the network is trained at different resolutions.
Such a process is known to be very efficient in other fields, both from a computational point of view, and from skipping local minima~\cite{Modersitzki2004,HabModMG04,WarnerEtAt2013}.
The method also allows for the classification of low-resolution images by networks that have been designed for and trained using high-resolution images {\em without} interpolation of the coarse scale image to finer scales.

We also present a method for scaling the number of layers in CNNs. Our method is based on the interpretation of the forward propagation in CNN as a discretization of a time-dependent nonlinear differential equation. In that framework the number of layers corresponds to the number time steps. Our observation motivates the use of  multi-level learning algorithms that accelerate the training of deep networks by solving a series of learning problems from shallow to deep architectures. 

The rest of this paper is structured as follows. In the next section, we show the connection between time dependent differential equations and CNNs. This connection allows us to introduce the optimization problem as a dynamic control problem. In Sec.~\ref{sec3} we present multiscale methods connecting CNNs across image resolutions and depths.  In Sec.~\ref{sec4} we demonstrate the potential of our methods using image classification benchmarks. Finally, in Sec.~\ref{sec5} we summarize the paper.

\section{An Optimal Control Perspective on CNNs} 
\label{sec:optControl}

In this section we derive a fully continuous formulation of deep convolution neural networks in the framework of optimal control. 
In Sec.~\ref{sub:fwd} we give a continuous interpretation of the spatial convolution and the forward propagation. In Sec.~\ref{sub:optContr} we discuss the remaining components of CNNs and present the continuous optimal control problem.

We focus on image classification and assume we are given training data consisting of discrete $d$-dimensional images $\bfx^{(1)}, \bfx^{(2)}, \ldots, \bfx^{(m)} \in \R^n$ and corresponding labels $\bfc^{(1)}, \bfc^{(2)}, \ldots, \bfc^{(m)} \in \R^\ell$.
In this paper, we consider $d = 2$ and thus $n$ corresponds to the number of pixels in the data. 
As common in image processing, we interpret the image data as discretization of a continuous function $x : \Omega \to \R$ at the cell-centers of a rectangular grid with $n$ equally sized pixels. Here, $\Omega\subset\R^d$ denotes the image domain and for simplicity we assume square pixels of edge length $h>0$.  We denote the number of layers in the deep CNN by $N$.

\subsection{Forward Propagation as a Nonlinear Differential Equation} 
\label{sub:fwd}

In this section we establish the interpretation of the forward propagation Residual Neural Networks (ResNN)~\cite{he2016identity} as a nonlinear differential equation. A simple way to write the forward propagation of a discrete image $\bfx\in\R^n$ through a  ResNet is
\begin{eqnarray}
\label{cnn}
\bfy_{k+1} = \bfy_k + \delta t F(\bfy_{k},\bftheta_k), \quad \quad \bfy_0 = \bfL \bfx, \quad \forall k=0,1,\ldots,N.
\end{eqnarray}
 Here, $\bfy_0 \in \R^{n_f}$ are the input features,  $\bfy_1,\ldots,\bfy_N$ are the hidden layers, and $\bfy_{N+1}$ are the output layers. The matrix $\bfL$ maps the input image into the feature space $\R^{n_f}$. This matrix can be "learned" or fixed.
 The parameters
 $\bftheta_k $ need to be determined by the "learning" process.
We generalize the original ResNet model by adding the parameter $\delta t>0$, which helps to derive the continuous interpretation below (the original formulation is obtained for $\delta t = 1$). 

 A common choice in CNN is to have the function $F$
 as a convolution with parameter $\bftheta$ that represent the
 convolution weights and bias, leading to the explicit
 expression
 \begin{equation}
 \label{conv}
 F(\bfy,\bfs,\bfb)=  \sigma_{\alpha}\left( \bfK(\bfs) \bfy + \bfb \right).
 \end{equation}
Here $\bfK(\bfs)$ is a convolution matrix, which is a circulant matrix that represents the convolution and depends on the stencil  or convolution kernel, $\bfs\in\R^{n_s}$, $\bfb \in \R^N$ is a bias vector and $\sigma_{\alpha}$ is an activation function.
The size of the stencil is typically much smaller than the number of pixels in the image and thus $\bfK(\bfs)$ is a sparse matrix. 
Next, we interpret the depth of the network in a continuous framework.
We start by rewriting the forward propagation~\eqref{cnn} as
\begin{eqnarray}
\label{cnn1}
\frac{\bfy_{k+1} - \bfy_k}{\delta t}=   \sigma_{\alpha}(\bfK(\bfs_k) \bfy_{k} + \bfb_k).
\end{eqnarray}
The left hand side of the above equation is a finite difference approximation to the differential operator $\partial_t \bfy$ with step size $\delta t$. While the approximation used in the original ResNet (using $\delta t = 1$) is valid if features change sufficiently slowly, the obtained dynamical system can be chaotic if the features change quickly.
Having a chaotic system as a forward problem implies that one can expect difficulties when considering the learning problem. Therefore, our first goal is to stabilize the forward propagation process.

To obtain a fully continuous formulation of the forward propagation we note that the convolution weights $\bfs$ can be seen as a discretization of continuous functions $s: \Omega \to \R$ (whose support is limited to a small region around the origin). This allows to interpret $\bfK(\bfs) \bfy$ as a discretization of $s *  y$.
Upon taking the limit $\delta t \to 0$ in~\eqref{cnn1} we obtain the {\em continuous} forward propagation process
\begin{equation}
 \label{cnncont}
\frac{d y}{d t}(t) = \sigma_{\alpha}\left(  s(t) * y(t) + b(t) \right), \quad y(0) = L x,
\end{equation}
for all $t \in [0,T]$, where $T$ is the final time corresponding with the output layer.
General stability of ordinary differential equations (ODEs) applies for this process and in particular, it is easy to verify that the system of ODEs is stable as long as the real part of the eigenvalues of the convolution are non-positive.
A second well known problem is vanishing gradients~\cite{BengioEtAl1994}. This implies that the eigenvalues of the convolution have a strong negative real part. Note that if the eigenvalues of $\bfK$ are imaginary, then no decay of the signal occurs, and no vanishing gradients are expected. This can aid in choosing and initializing the network parameters.

Given the continuous forward propagation in~\eqref{cnncont} we interpret~\eqref{cnn1} as a forward Euler discretization with a fixed time step size of $\delta t$. Thus, the forward propagation is stable as long as the real parts of the eigenvalues of the convolution and the time steps are sufficiently small.  We note that there are numerous methods for time integration, some of which provide superior stability of the forward propagation.


\subsection{Optimal Control Formulation of Supervised Learning} 
\label{sub:optContr}
Having discussed forward propagation, we now briefly review classification and give continuous and discrete formulations of the learning problem.

The hypothesis or classification function, which predicts the label for each data using the values at the output layer, $\bfy_{N+1}$, can be written as
\begin{eqnarray}
\label{classifier}
\bfc_{\rm pred} = g(h^d \bfW^\top\bfy_{N+1} + \mu), 
\end{eqnarray}
where the columns of $\bfW \in \R^{n_f \times \ell}$ are classification weights and $\mu \in \R^\ell$ are biases  for the respective classes.
Commonly used choices are softmax, least-squares, logistic regression, or support vector machines.
We have generalized the common notation by adding the parameter $h^d$ that allows to interpret $\bfW^\top \bfy_{N+1}$ as a midpoint rule applied to the standard $L_2$ inner product $(w_j,y)_{L_2} = \int_{\Omega} w_j(r) y(r) dr$ for a sufficiently regular function $w_j : \Omega \to \R$. The $j$th column of $\bfW$ is the discretization of $w_j$ at the cell-centers of the grid. This generalization allows to adjust the weights across image resolutions.

For training data containing of continuous functions $x^{(1)}, \ldots, x^{(m)}$ and labels $\bfc^{(1)}, \ldots, \bfc^{(m)}$, learning consists of solving the optimal control problem
\begin{subequations}
\begin{eqnarray}
\min_{w, \mu,s, b} &  \frac{1}{m} \sum_{j=1}^m S(g((w,y^{(j)}(T))_{L_2} + \mu) ,\bfc^{(j)}) + R(w,\mu,s, b)\\
{\rm subject\ to} &
\frac{d y^{(j)}}{d t}(t) = \sigma_{\alpha}\left(  s(t) * y^{(j)}(t) + b(t) \right), \quad y^{(j)}(0) = L x^{(j)}, \quad \forall j=1,\ldots,m.
\end{eqnarray}
\end{subequations}
Here $S$ is a loss function measuring the mismatch between the predicted and known label and $R$ is a regularization function that penalizes undesired features in the parameters and avoids overfitting.
Typically, the problem is not solved to a high accuracy and low accuracy solutions are sufficient. A validation set is often used to determine the stopping criteria for the optimization algorithm.
 
For completeness we note that a discrete version of the optimal control problem is 
\begin{subequations}
\label{opt}
\begin{eqnarray}
\min_{\bfW, \mu,\bfs_{1,2,\ldots,N}, b_{1,2,\ldots,L}} &  \frac{1}{m} \sum_{j=1}^m S(g(h^d \bfW^\top\bfy_{N+1}^{(j)} + \mu) ,\bfc^{(j)}) + R(\bfW,\mu,\bfs_{1,2,\ldots,N}, b_{1,2,\ldots,N})\\
{\rm subject\ to} &
\bfy^{(j)}_{k+1} = \bfy^{(j)}_k + \delta t \sigma_a (\bfy^{(j)}_{k},\bftheta_k), \quad \quad \bfy^{(j)}_0 = \bfL \bfx^{(j)}, \quad \forall j=1,\ldots,m.
\end{eqnarray}
\end{subequations}

Note that for simplicity we have ignored the pooling layer, although it can be added in general (The necessity of pooling has been debated in~\cite{DBLP:journals/corr/SpringenbergDBR14}).

\section{Multiscale Methods}\label{sec3}

In this section we present new methods for scaling deep CNNs along two dimensions. In Sec.~\ref{sub:restriction} and Sec.~\ref{sub:prolongation} we discuss restriction and prolongation of convolution operators as a way to scale CNNs along image resolution. In Sec.~\ref{sub:timeProlongation} we scale the depth of the network to simplify initialization and accelerate training.

\subsection{From High-Resolution to Low-Resolution:  Restricting Convolution Operators} \label{sub:restriction}

Assume first that we are given some image data, $\bfy_h$, on a mesh with pixel size $h$ and a stencil, $\bfs_h$ that operates on this image. Assume also that we would like to apply the fine mesh convolution to an image, $\bfy_H$, given on a coarser mesh with pixel size $H > h$. In other words the goal is to find a stencil $\bfs_H$ for which the coarse mesh convolution is equivalent to refining the image data and applying fine mesh convolution with $\bfs_h$. This problem is well studied in the multigrid literature~\cite{tos}.

Our method for restricting the stencil follows the algebraic multigrid approach; see, e.g.,~\cite{tos} for details and alternative approaches using re-discretization. We assume that the following connection holds between the fine mesh image, $\bfy_h$, and the coarse mesh image $\bfy_H$
\begin{eqnarray}\label{pro-rest}
\bfy_H = \bfR \bfy_h \quad {\rm and} \quad \tilde{\bfy}_h = \bfP \bfy_H.
\end{eqnarray}
Here, $\bfP$ is a prolongation matrix and $\bfR$ is a restriction matrix.
$\tilde{\bfy}_h$ is an interpolated coarse scale image on the fine mesh
and typically, $\bfR \bfP = \gamma\bfI$ for some $\gamma$ that depends on the dimensionality of the problem. The interpretation is that the coarse scale image is obtained using some linear transformation from the fine scale image (e.g. averaging). Conversely, an approximate fine scale image can be obtained from the coarse scale image by interpolation. This interpretation can easily be extended to 3D data to allow, e.g., classification of videos.

Let $\bfK_h(\bfs_h)$ be the sparse matrix that represents the convolution on the fine scale. This matrix operates on a vectorized image and is equivalent to convolving the vector $\bfy_h$ with the stencil, $\bfs_h$. The matrix is circulant and sparse with a few non-zero diagonals. Our goal is to build a coarse scale convolution, $\bfK_H$ that operates on a vector $\bfy_H$
and is consistent with the operation of $\bfK_h$ on a fine scale vector $\bfy_h$.  Using the prolongation and restriction we obtain that
\begin{equation}
\label{KH1}
\bfK_H \bfy_H = \bfR \bfK_h \bfP \bfy_H.
\end{equation}
That is, given $\bfy_H$ we first prolong it to the mesh $h$, then operate
on it with the matrix $\bfK_h$, which yields a vector on mesh $h$. Finally, we restrict the result to the mesh $H$. This implies that the coarse scale convolution matrix can be expressed as
$\bfK_H = \bfR \bfK_h \bfP$.
This definition of the operator can be built directly using any interpolation/restriction operators. Furthermore, assuming that the stencil, $\bfs_H$ is constant on the coarse mesh (that is, it is not changing on the mesh as commonly assumed in CNN), it is straightforward to evaluate it without generating the matrix $\bfK_H$, as commonly done in algebraic multigrid.

\begin{example}
	To demonstrate the concept of how to the convolution changes with resolution,
	we use the following simple example demonstrated in Figure~\ref{fig1}.
	We select an image from the MNIST data set (bottom left) and convolve it
	with the fine mesh convolution parameterized by the stencil
	$$ \bfs_h = \begin{pmatrix}
	-0.89 & -2.03 & 4.30 \\
	-2.07 & 0.00 & -2.07 \\
	4.39 & -2.03 & 1.28
	\end{pmatrix} $$
	obtaining the image in the bottom right panel of Figure~\ref{fig1}. Now, by restricting the weights using the algebraic multigrid approach, we obtain that on a coarse mesh the weights are,
	$$ \bfs_H = \begin{pmatrix}
	-0.48 & -0.17 & 0.82 \\
	-0.15 & -0.80 & 0.37 \\
	0.84 & 0.40 & 0.07
	\end{pmatrix}. $$
	These weights are used on the coarse scale image (top left panel of Figure~\ref{fig1}) to construct the filtered image on the top right panel of Figure~\ref{fig1}.
	\begin{figure}
	  \begin{center}
	    \includegraphics[width=.35\textwidth]{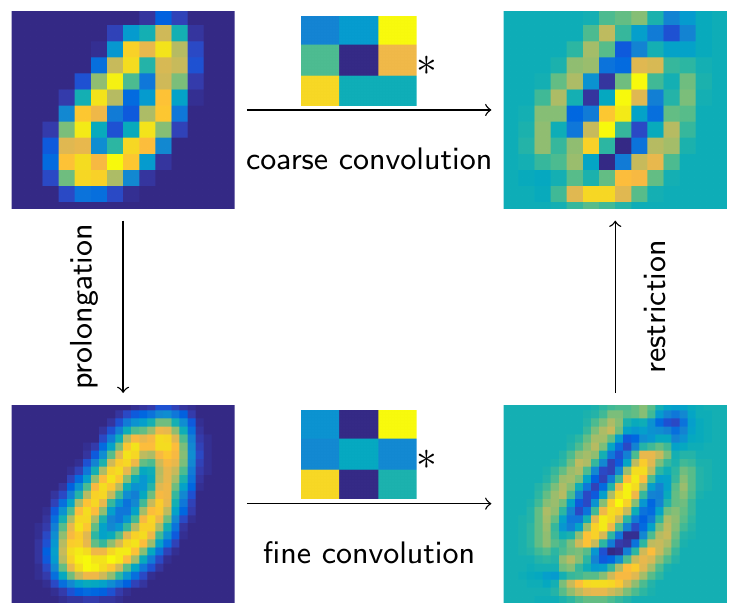}
	  \end{center}
	  \caption{Illustration of a fine mesh vs. coarse mesh convolution. \label{fig1}}
	\end{figure}
	Looking at the weights obtained on the coarse mesh, it is evident that they are significantly different from the fine scale weights. It is also evident from the fine and coarse images, that adjusting the weights on the coarse mesh is necessary if we are to keep a faithful transformation between the images.
\end{example}

The interpretation of images and convolution weights as continuous functions, allows us to work with different image-resolutions. This has two important consequences. First, assume that we have trained our network on some fine scale images and that we are given a coarse scale image. Rather than interpolating the image to a fine mesh (which can be memory intensive and computationally expensive), we transform the {\bf stencils} to a coarse mesh and use the coarse mesh stencils to classify the coarse scale image. Such a process can be particularly efficient when considering the classification of videos on mobile devices where expanding the video to high resolution can be computationally prohibitive. A second consequence is that we are able to train the network on a coarse mesh and then interpolate the result to a fine mesh. As we see next, this allows us to use a process of image pyramid or multi-resolution for the solution of the optimization problem that is at the heart of the training process.

\subsection{From Low-Resolution to High-Resolution: Prolongating Convolution Operators}
\label{sub:prolongation}

Understanding how to move between different scales allows us to construct efficient algorithms that use inexpensive coarse mesh representations of the problem in order to initialize the problem on finer scales. This is similar to the classical image pyramid process~\cite{HabModMG04} and multilevel methods that are used in applications that range from full waveform inversion to shape from shading~\cite{ZTCM99}. The idea is to solve the optimization problem on a coarse mesh first in order to initialize fine grid parameters.
The algorithm is summarized in Algorithm \ref{alg1}.
\begin{algorithm}[t]
\begin{algorithmic}[1]
\STATE{ Restrict the images $n_c$ times}
\STATE{ Initialize the stencils, biases, and classification weights}
\FOR{$i=n_c:-1:1$}
\STATE{Solve the optimization problem \eqref{opt} on mesh $i$  from its initial point}
\STATE{Prolong the stencils to level $i-1$}
\STATE{Update the classifier weights}
\ENDFOR
 \end{algorithmic}
\caption{ \label{alg1} Multigrid Prolongation}
\end{algorithm}

Solving each optimization problem on coarser meshes is  cheaper than solving the problem on finer meshes. In fact, when an image is coarsened by a factor of 2, each convolution step is 4 times cheaper in 2D and 8 times cheaper in 3D. In some cases, such a process leads to linear complexity of the problem~\cite{tos}.

In order to apply such algorithms in our context, we need to address the transformation of the coarse scale operator to a fine scale one. This process is different than classical multigrid where the operator on a fine mesh is given, and a coarse scale representation is desired. As previously discussed, we use the classical multigrid result to transform a fine mesh operator to a coarse one
\begin{eqnarray}
\label{KhH}
\bfK_H  = \bfR \bfK_h \bfP.
\end{eqnarray}
In the classical multigrid implementation, one has a hold on the {\em fine scale} operator $\bfK_h$, and the goal is to compute the coarse scale operator $\bfK_H$. In our application, throughout the mesh continuation method, we are given the {\em coarse mesh} operator and we are to compute the fine mesh operator. In principle, there is no unique fine scale operator given a coarse scale one; however, assuming that the fine scale operator is a convolution with fixed stencil (as in the case of CNN), there is a unique solution. This is a classical result in Fourier analysis on multigrid methods~\cite{tos}. Since  \eqref{KhH} represents a linear connection between $\bfK_h$ and $\bfK_H$, we extract $n_K^2$ equations (where $n_K$ is the size of each convolution stencil)
that connect the fine scale convolutions to the coarse scale ones. For a convolution stencil of size $3^2$ and linear prolongation/restriction, this is a simple $9 \times 9$ linear system that can be easily solved to obtain the fine scale convolution. A classical multigrid result is that this linear system is well-posed. Assuming that the coarse mesh is a $n_K \times n_K$ stencil and that
the interpolation is linear, the fine mesh stencil is also a $n_K \times n_K$
stencil which is uniquely determined from the coarse mesh stencil.

\subsection{From Shallow to Deep Networks} 
\label{sub:timeProlongation}
In this section we consider scaling the number of layers in the network as another way to use the continuous framework. 
In our case we gradually increase the number of layers keeping the final time $T$ constant in order to accelerate learning by re-using parameters from shallow networks to initialize the learning problem for the deeper architecture. Note that the number of layers in the network corresponds to the number of discretization points in the discrete forward propagation. Similar ideas have been used in multigrid~\cite{BornemannDeuflhard1996} and image processing; see, e.g.,~\cite{ModSiamBook}. 

To solve a learning task in practice, we first solve the learning problem using a network with only a few layers. Subsequently, we prolongate the estimated parameters of the forward propagation to initialize the optimization problem for the next network that features, e.g., twice as many layers. To this end, simple linear interpolation can be used. We repeat this process until the desired network depth is reached.

Besides realizing some obvious computational savings on shallower networks, the main motivation behind our approach is to obtain good starting guesses for the next level. This is key since, while deeper architectures offer more flexibility to model complicated data-label relation, deeper networks are notoriously difficult to initialize. Another advantage when using second-order learning algorithms is the faster convergence rate obtained by warm starting.

\section{Experiments}\label{sec4}

In this section, we demonstrate the benefits of the proposed multiscale algorithms for CNN using two supervised image classification problems. 

\subsection{Classification of Images Across Resolutions}\label{sub:exp1}

We demonstrate that using the continuous formulation we are able to classify low-resolution images using CNNs trained on high-resolution images and vice versa. Here, no additional learning is performed and, for example, classifying the low-resolution images does require neither interpolation nor high-resolution convolutions. This is important for efficient classification on mobile devices etc.

We consider the MNIST dataset and independently train two networks with two layers each using the coarse and fine data, respectively. The MNIST dataset that consists of 60,000 labeled images each with $28\times 28$ pixels. Since the images are rather coarse, we use only two levels. To obtain coarse scale images the fine scale images are convolved with a Gaussian, and restricted to a coarse mesh using the operator introduced above. This yields a coarse mesh data consisting of $14\times 14$ images. We randomly divide the datasets into a training set consisting of $50,000$ images, and a validation set consisting of $10,000$ images.
In all experiments, we choose a CNN with identical layers,  $\tanh$ activation function, and a softmax classifier. For optimization we use a Block-Coordinate-Descent (BCD) method. Each iteration consists of one Gauss-Newton step with sub-sampled Hessian to update the forward propagation parameters and 5 Newton steps to update the weights and biases of the classifier. To avoid overfitting and stabilize the process we enforce spatial smoothness of the classification weights and smoothness across layers for the propagation parameters through derivative-based regularization.

The validation accuracy for the networks on the resolutions they are trained on is around 98.28\% and 98.18\% for the coarse and fine scale network, respectively. Next, we prolongate the classification weights and apply the multigrid prolongation to the convolution kernels from the coarse network to the fine resolution. Using only the results from the coarse level and no training on the fine level, we get a validation accuracy of $91.02$\%. For comparison, using the original convolution kernels gives a validation accuracy of $61.02$\%.

Next, we restrict the classification weights and convolution kernel of the network trained on fine data. Using this, gives a validation accuracy $94.92$\% (compared to $84.09$\% without restricting the kernels). Again, we note that no training is performed on the coarse resolution.

\subsection{Shallow to Deep Training}\label{sub:exp2}

We show the benefit of the multilevel training strategy using the MNIST example. We solve a sequence of training problems for CNNs whose depths increase in powers of two from 2 layers up to 64. For each CNN, we estimate the parameters using 20 iterations of the BCD. Except the number of layers, all parameters are chosen identical to the previous experiment. 

\begin{figure}[t]
	\begin{center}
		\includegraphics[width=.9\textwidth]{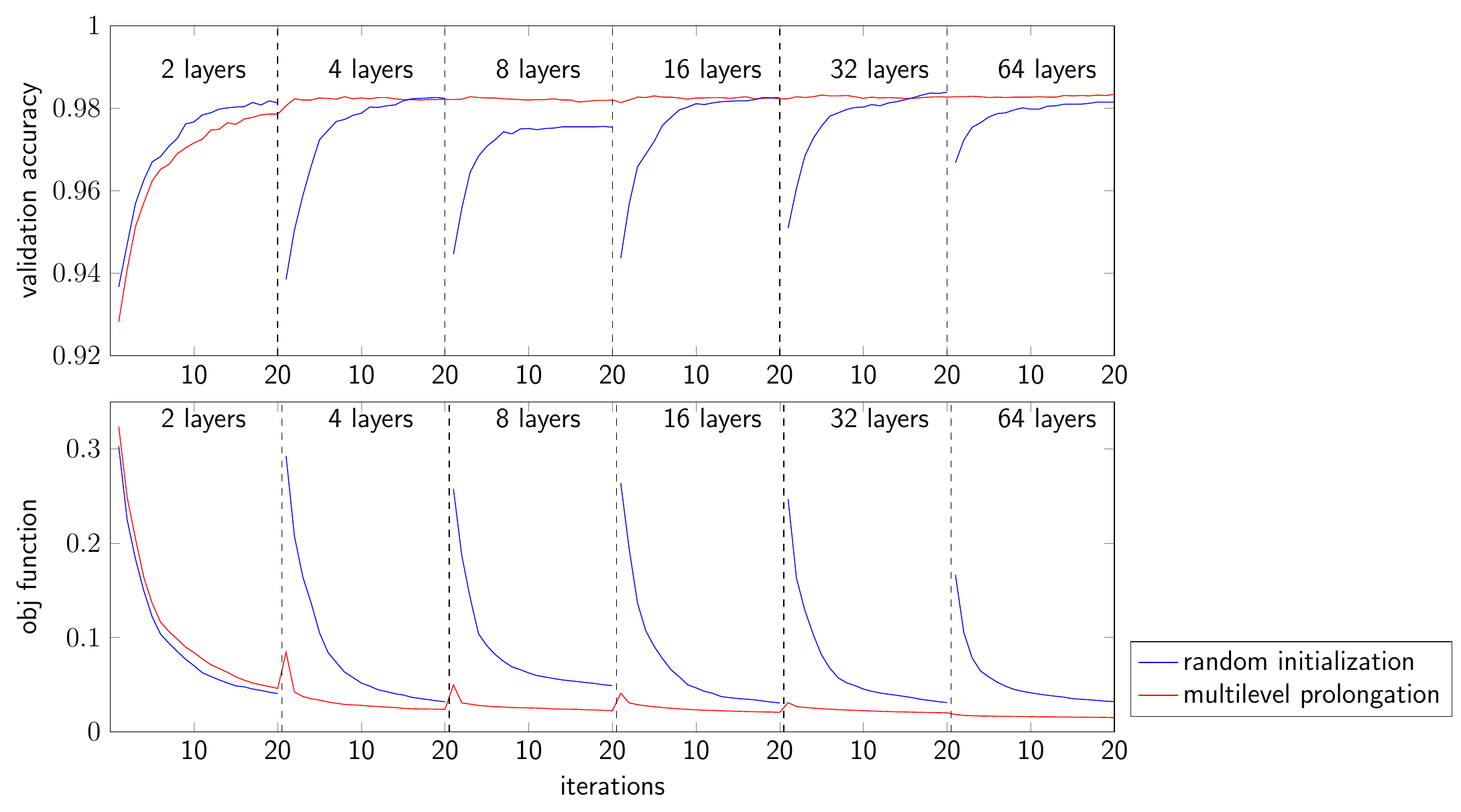}
	\end{center}
	\caption{Multilevel convergence for MNIST problem.}
	\label{fig:mlconv}
\end{figure}

We compare the convergence properties of the learning algorithm using random initialization and the proposed multiscale initialization, which uses the prolongated network parameters from 
the previous level. The validation accuracy and the value of the loss function can be seen in~Fig.\eqref{fig:mlconv}. 
It can be seen that the initial guesses provided by the multiscale process have a lower value of the loss function and higher validation accuracy. For the deeper networks where training is most costly, the optimal accuracy is reached after only a few iterations using the multiscale method while for random initialization more iterations are needed to achieve a comparable accuracy.

\subsection{Multiscale CNN Training on ImageNet}\label{sub:exp3}

\begin{figure}[t!]
	\centering
	\begin{tabular}{@{}c@{}c@{}}
		\includegraphics[width=0.5\textwidth]{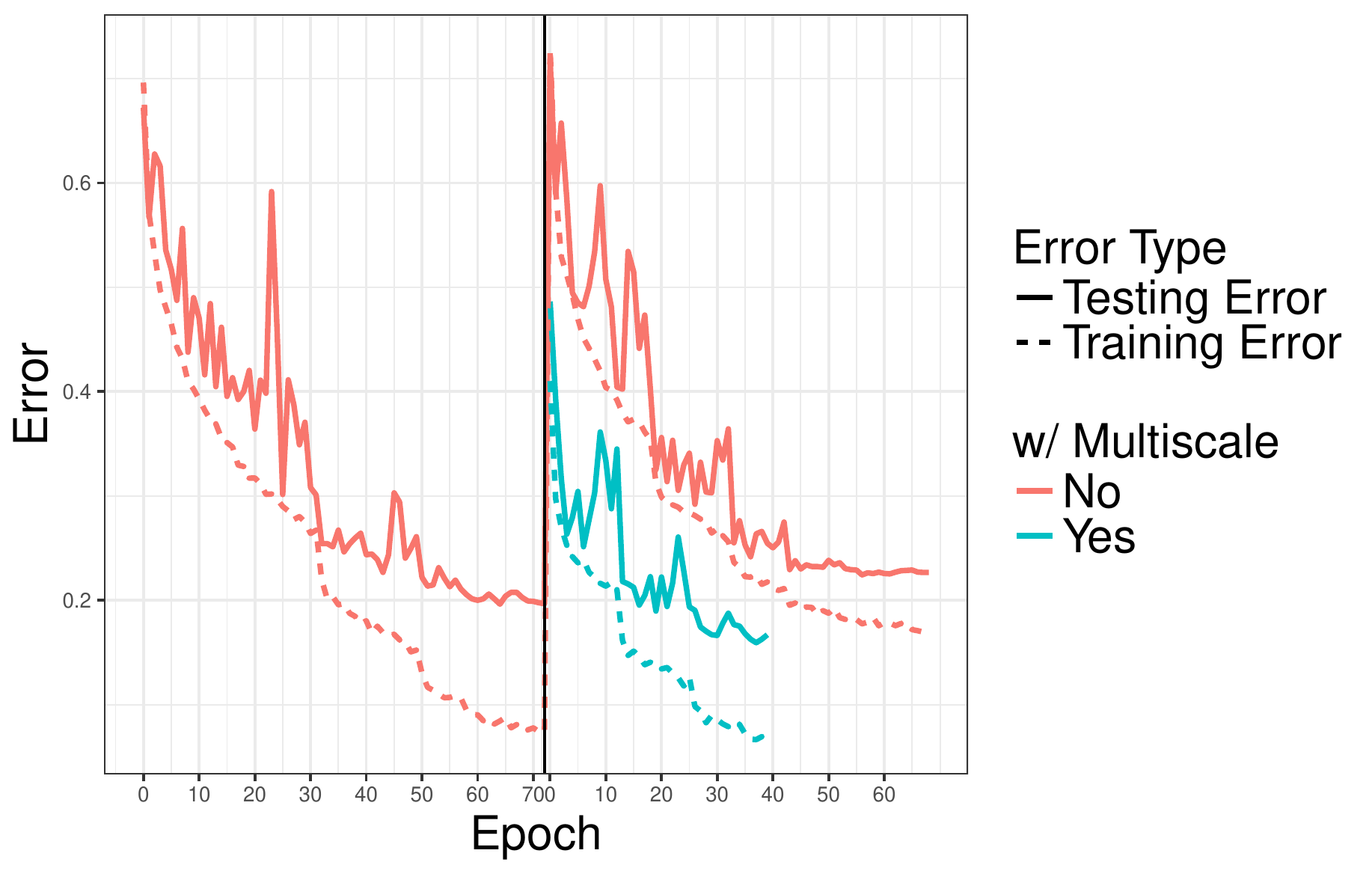} &
		\includegraphics[width=0.5\textwidth]{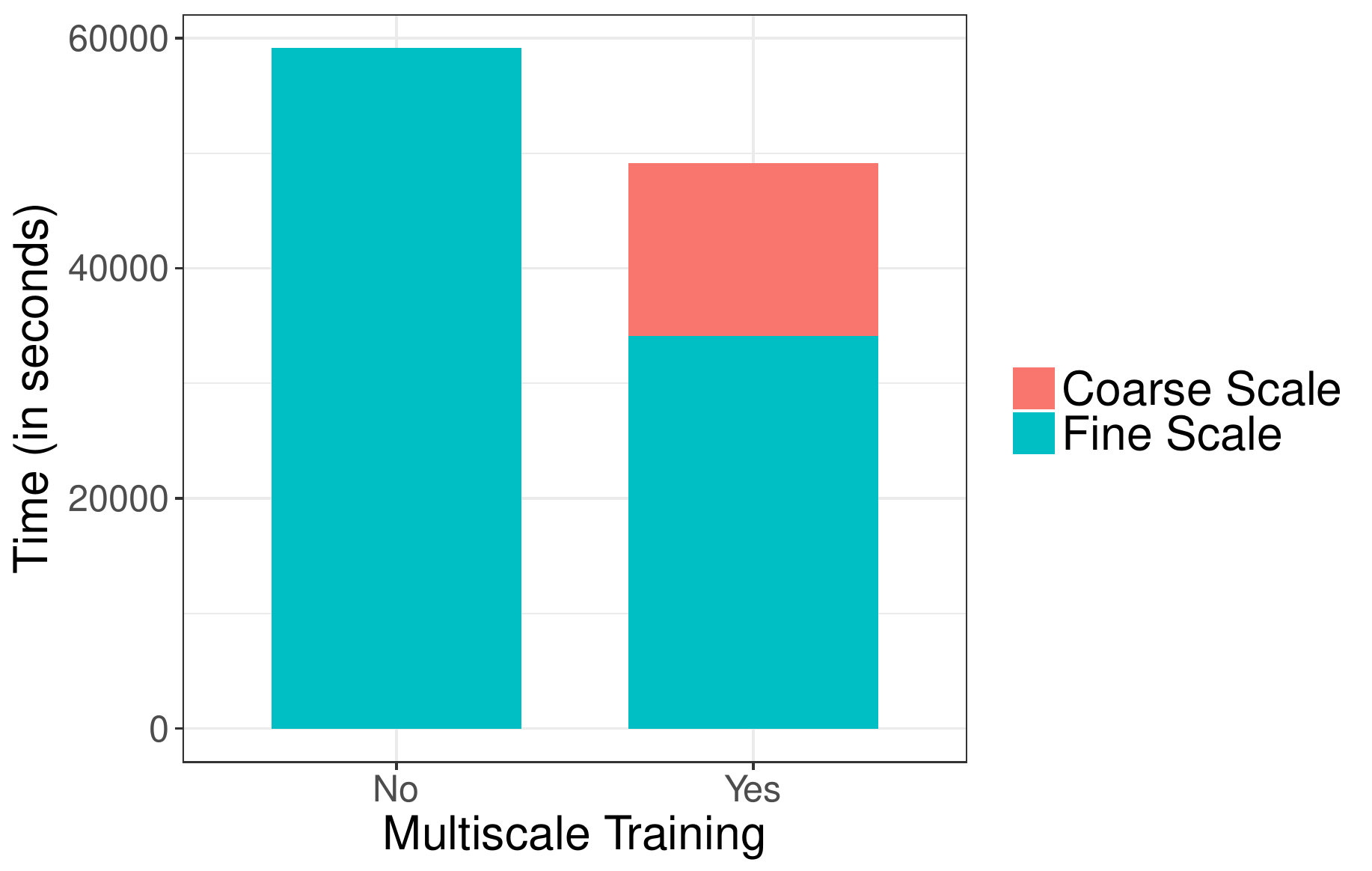} \\
	\end{tabular}
	\caption{Results on ImageNet-10. Left: Comparison of training and validation accuracy for fine mesh training using $224 \times 224$ images (red) and our coarse-to-fine multiscale approach using weights trained from $112 \times 112$ (blue). Note the multiscale method achieves superior training and validation accuracy in fewer epochs. Right:  Slight computational savings are achieved using our multiscale approach compared to independent learning on each resolution.  }
	\label{fig:imagenet10-results}
\end{figure}

We demonstrate the computational benefit of coarse mesh training compared with only training the fine mesh CNN. To this end, we select ten categories from the ImageNet dataset~\cite{ILSVRC15}. The ImageNet consists of images that are of varying dimensions and hence, we pre-process the images to be of dimension 224-by-224 as also proposed in~\cite{he2016deep}. On this resolution level, the image quality is sufficiently high to visually recognize the objects in the images and the discrete images appear smooth, i.e., are free of block artifacts. This choice leads to non-trivial training problem, where one can expect to reduce training time using our multigrid approach. For each category, there are 1,300 images for the total of 13,000 images. We randomly divide the data into 10,000 images used for training and 3,000 images used for validation. We use ResNet-34 architecture shown in Fig.~\ref{fig:imagenet10-results} of ~\cite{he2016deep} with two differences, first, the first CNN kernel is of dimension $3 \times 3 \times 64$ rather than $7 \times 7 \times 64$ and second, we did not use the fully connected layer, but rather, the output of the average pooling is connected to the classification layer with softmax activation directly. Note that the average pooling ensures the dimension of the penultimate layer to be identical regardless of the dimension of the input layer.  

\begin{figure}[t!]
	\centering
	\begin{tabular}{@{}c@{}c@{}}
		\includegraphics[width=0.45\textwidth]{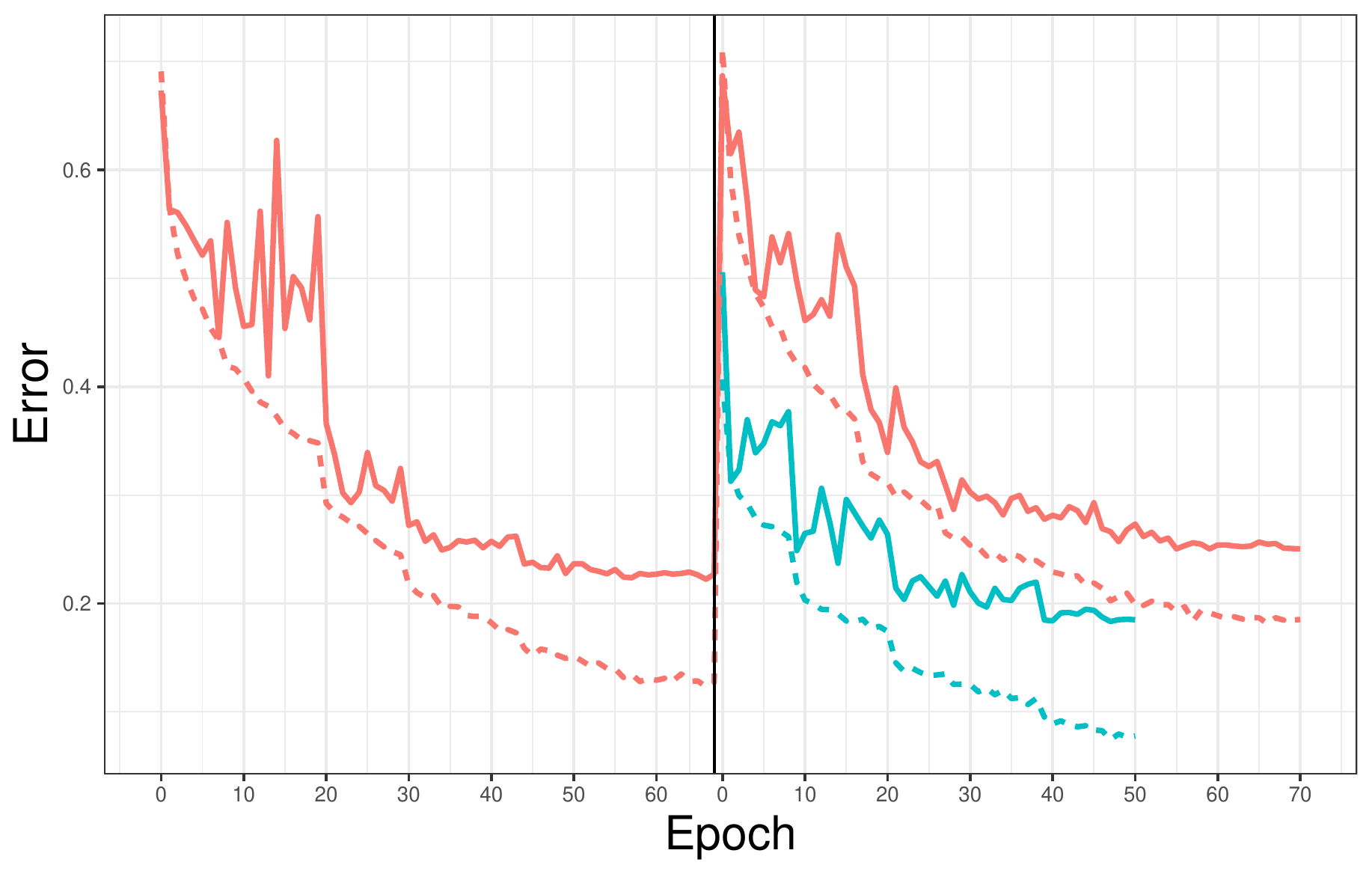} &
		\includegraphics[width=0.45\textwidth]{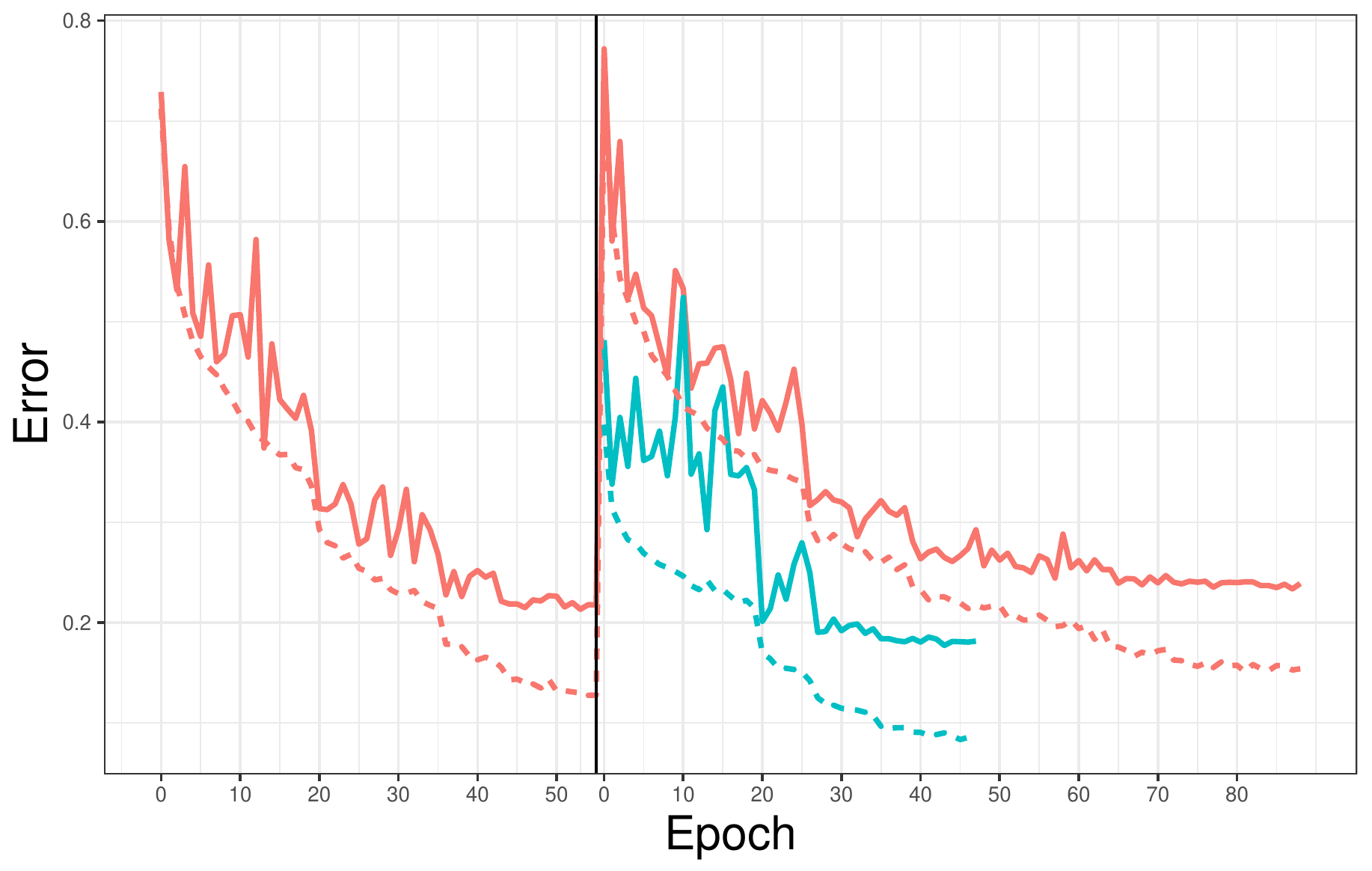} \\
		\includegraphics[width=0.45\textwidth]{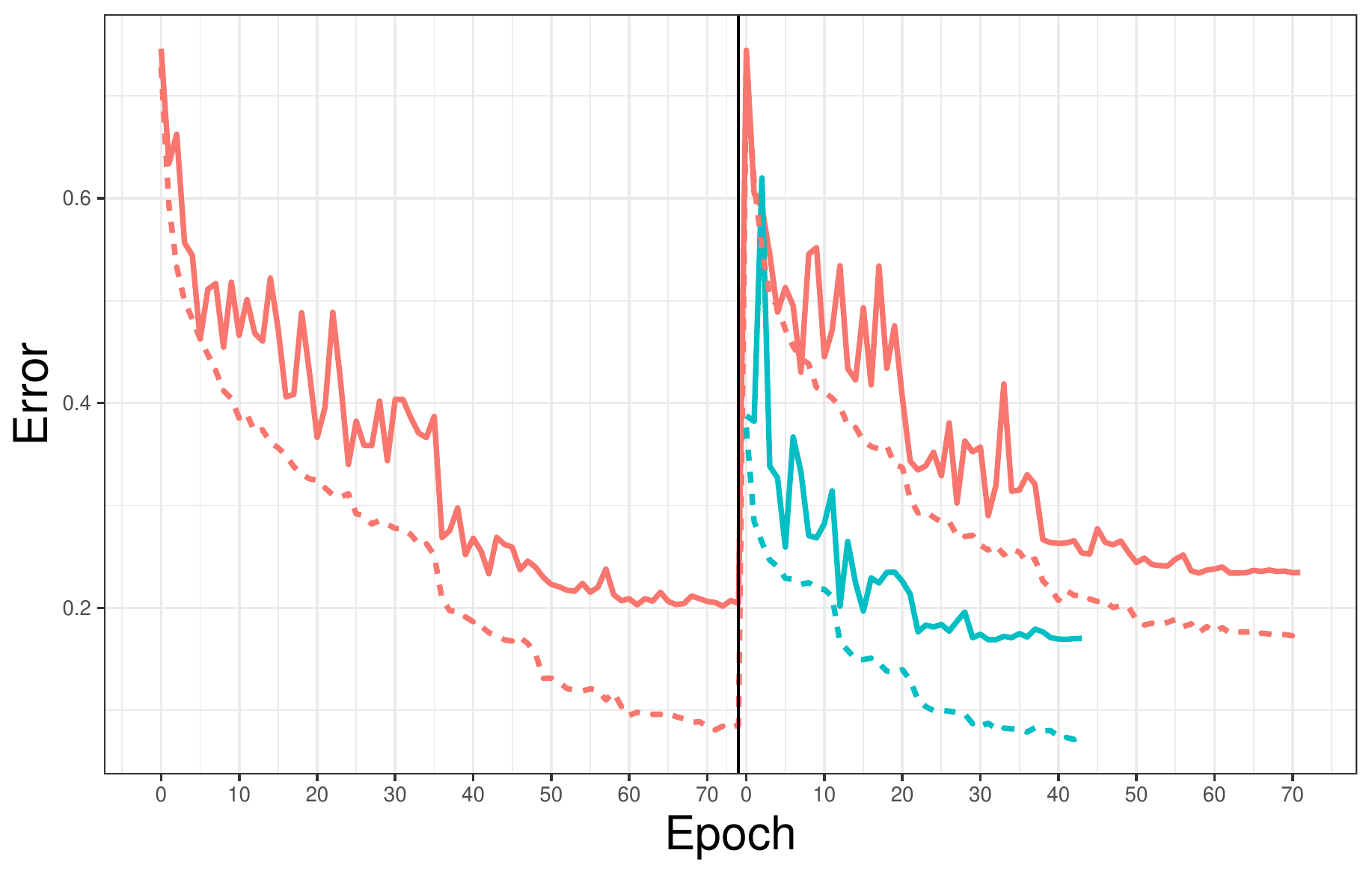} &
		\includegraphics[width=0.45\textwidth]{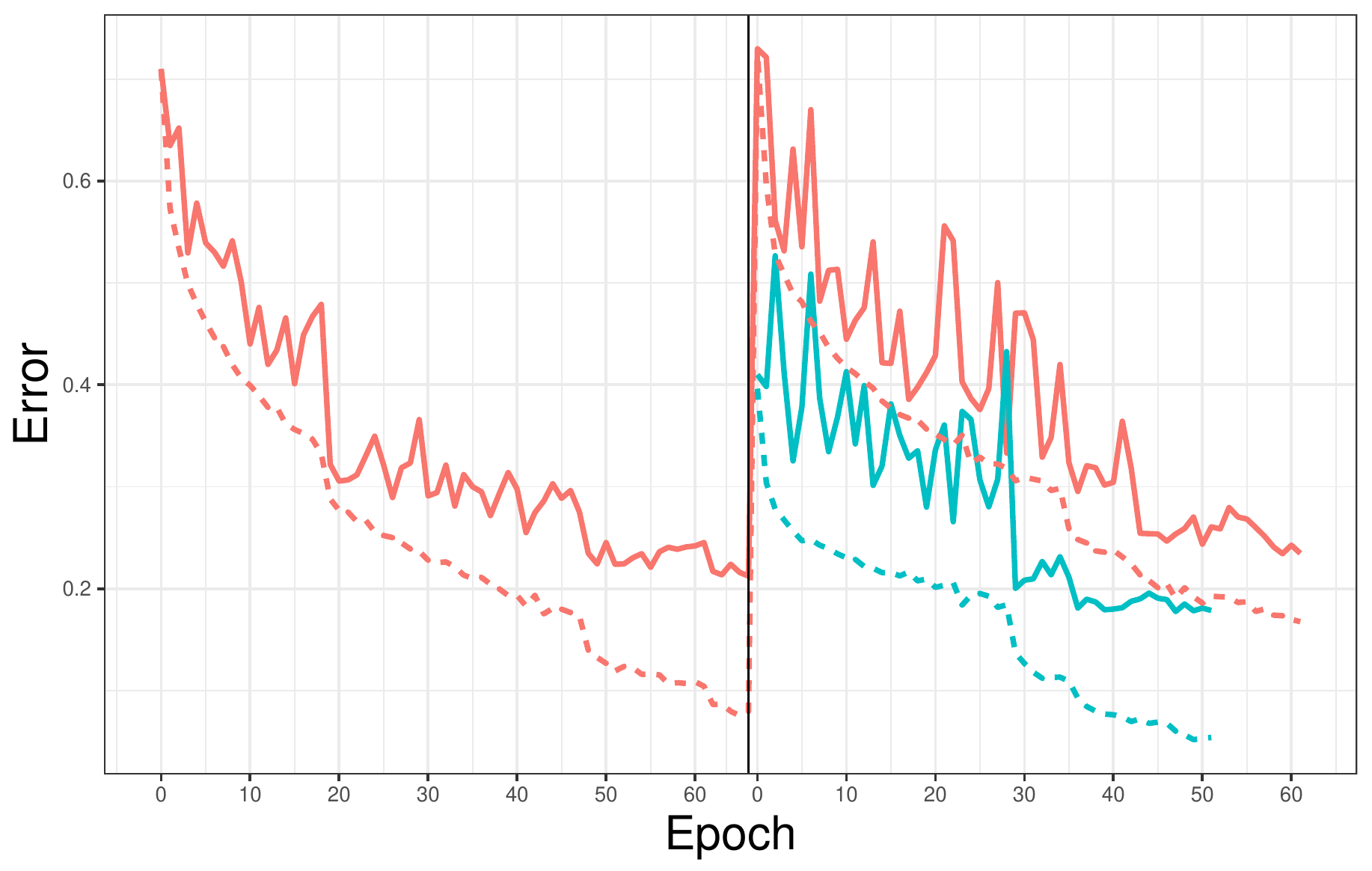} \\
	\end{tabular}
	\caption{Additional results on ImageNet-10. Refer to Figure~\ref{fig:imagenet10-results} (a) for the legend.}
	\label{fig:imagenet10-additional-results}
\end{figure}

We demonstrate the computational benefit of coarse mesh training compared with only training the fine mesh CNN. The first step is coarse mesh training. To this end, we restrict the $224 \times 224$ images to a $112 \times 112$ mesh and train ResNet-34 on the coarsened images. Then, we obtain the weights for the finer scale using the method described in Section~\ref{sub:prolongation}. For comparison, we also train ResNet-34 using the $224 \times 224$ image data directly. We show in the left subplot of Fig.~\ref{fig:imagenet10-results} that the multiscale approach results in considerable reduction in the number of epochs before reaching convergence. We adopt early stopping to stop training if the loss does not improve for 10 Epochs. In the right subplot of Fig.~\ref{fig:imagenet10-results}, we show the total time to training for multiscale, sum of the time to train for the coarse scale and time to train for the fine scale, compared to directly training on $224 \times 224$. Note that the multiscale approach not only requires fewer epochs and lower runtime, but also gives lower training and validation errors. To ensure that this is not a phenomenon that is specific to the given train-test data split, we report the average over 5 splits; we have provided additional plots in Figure~\ref{fig:imagenet10-additional-results}. In all train-test data splits, the multiscale approach produce lower training and validation error in smaller number of epochs.

\section{Conclusions}
\label{sec5}

In this work, we explore the connection between optimal control and 
training deep Convolution Neural Networks (CNNs) that enables learning across scales. 
The foundation of our approaches is a continuous image model and interpreting forward propagation in CNN as discretization of a time-dependent nonlinear differential equation. 
We showed how this mathematical framework can be used to scale deep CNNs along two dimensions: Image resolution and depth.  
While the obtained multiscale approaches are new in deep learning they are commonly used for the numerical solution of related optimal control problems, e.g., in image processing and parameter estimation.

Our method for connecting low- and high-resolution images is unique in that it scales the parameter of the network rather than interpolating the image data to different resolutions.
To this end, we present an algebraic multigrid approach to computing convolution operators that are consistent with coarse and fine scale images. We exemplified the benefit of our approach in two ways. In Sec.~\ref{sub:exp1}, we show that CNNs trained on fine resolution images can be adapted and used to classify coarse resolution images and vice versa.
Our method is advantageous when memory is limited and interpolation of images or videos is not practical.
In Sec.~\ref{sub:exp3}, we demonstrate that it is possible to use coarse representation of the images to learn the convolution kernels and – after prolongation – use the weights to classify high-resolution images.
 
Our example in Sec.~\ref{sub:exp2} shows that scaling the number of layers of the CNN can improve and accelerate the training of deep networks through initialization using results from shallow ones. In our experiment this drastically reduces the number of iterations for the deep networks where cost-per-iteration is high.

Casting CNNs as a continuous optimal control problem of differential equations provides new insights into the field and motivates new ways to solve and regularize the learning problem.

\section{Acknowledgements} 
\label{sec:acknowledgements}
This work is supported in part by the US National Science Foundation (NSF) award DMS 1522599.


\end{document}